# Bounding Search Space Size via (Hyper)tree Decompositions

**Lars Otten** and **Rina Dechter**
Bren School of Information and Computer Sciences
University of California, Irvine, CA 92697-3435, U.S.A.
{lotten,dechter}@ics.uci.edu

## Abstract

This paper develops a measure for bounding the performance of AND/OR search algorithms for solving a variety of queries over graphical models. We show how drawing a connection to the recent notion of hypertree decompositions allows to exploit determinism in the problem specification and produce tighter bounds. We demonstrate on a variety of practical problem instances that we are often able to improve upon existing bounds by several orders of magnitude.

## 1 INTRODUCTION

This paper develops a measure for bounding the performance of search algorithms for solving a variety of queries over graphical models. It has been known for a while that the complexity of inference algorithms (e.g., join-tree clustering, variable elimination) is exponentially bounded by the tree width of the graphical model's underlying graph. The base of the exponent is often taken to be the maximum domain size.

More accurate bounds were derived by looking at the respective domain sizes and their product in each cluster in of tree decomposition of the underlying graph [Kjærulff, 1990]. These tighter bounds were used in selecting good variable orderings, for example. It was recently shown that these bounds are also applicable to search algorithms that explore the context-minimal AND/OR search graph [Dechter and Mateescu, 2007].

The shortcoming of these bounds is that they are completely blind to context-sensitivity hidden in the functions of the graphical model and especially determinism. When a problem possesses high levels of determinism, its tree width bound can be large while its search space can be extremely pruned, due to propagation of inconsistencies across functions.

Part of this shortcoming in worst-case complexity bounds is addressed by the more recent concept of hypertree decompositions [Gottlob et al., 2000]. It was shown that the maximum number of functions in the clusters of a hypertree decomposition (the hypertree width) exponentially bounds the problem complexity for constraint inference, a result that was extended to general graphical model inference in [Kask et al., 2005]. The base of the exponent in this case is the relation tightness, thus allowing the notion of determinism to play a role. However, in practice this bound often turns out to be far worse than the tree width bound, unless the problem exhibits substantial determinism [Dechter et al., 2008].

The contribution of this paper is in combining both ideas to tighten the existing bounds, using the relationship between AND/OR graph search and tree decompositions. Starting with the tree width bound, we show that one can also incorporate the concept of hypertree decompositions by greedily covering variables with tight functions. This yields better bounds on the number of nodes in the search graph, which translates directly to search complexity.

Tighter bounds are desirable for a number of reasons:

1. We can better predict parameters of the algorithm ahead of time (primarily the variable ordering for search), fitting the algorithm to the problem.

2. It enables us to dynamically update parameters during search, e.g., for dynamic variable orderings.

3. In a distributed setup, search can often be implemented as centralized conditioning followed by independent solving of the conditioned subproblems on different machines. Better bounds can help in balancing these two phases by varying the size of the central conditioning set [Silberstein et al., 2006].

We provide extensive empirical results on 112 probabilistic problem instances and 30 weighted constraint satisfaction problems. We show that exploiting determinism has a significant effect for a number of problem classes. We furthermore compare our bound to the exact size of the search space on a subset of feasible instances, and show that it can be very tight in some cases.

An approach that is related but orthogonal to the work here is described in [Zabiyaka and Darwiche, 2006], where the standard complexity measure of tree width is refined by taking into account functional dependencies – i.e., knowing one set of variables determines the values of another set. [Fishelson et al., 2005] develops a bound specifically for an interleaved variable elimination and conditioning algorithm on linkage analysis problems.

Section 2 provides the background and definitions. Section 3 discusses hypertree decompositions and related complexity bounds. In Section 4 we introduce our new bounding scheme, for which Section 5 provides empirical evaluation. Section 6 concludes.

## 2 PRELIMINARIES AND DEFINITIONS

In the following we will assume a *graphical model* given as a set of variables $X = \{x_1, \ldots, x_n\}$, their finite domains $D = \{D_1, \ldots, D_n\}$, a set of functions $F = \{f_1, \ldots, f_m\}$, each of which is defined over a subset of $X$, and a combination operator (typically sum, product or join) over all functions. Together with an marginalization operator such as $\min_X$ and $\max_X$ we obtain a *reasoning problem*.

The special cases of reasoning tasks which we have in mind are belief networks, (weighted) constraint networks or mixed networks that combine both. The primary tasks over belief networks are belief updating and finding the most probable explanation. They are often specified using conditional probability functions defined on each variable and its parents in a given directed acyclic graph (see Figure 1(a)), and use multiplication and summation or maximization as the combination and marginalization operators [Kask et al., 2005]. For constraint networks we are mainly concerned with problems of finding or enumerating solutions; they are defined using relations as functions, and relational join and projection as the combination and marginalization operators, respectively. For weighted constraint networks one typically has real-valued functions and summation and minimization as combination and marginalization operators, respectively.

### 2.1 EXPRESSING STRUCTURE

If one wants to analyze the complexity of a given problem instance, it has proven useful to look at the underlying structure of interactions between variables:

DEFINITION **2.1** *The **hypergraph** of a graphical model is a pair $H = (V, S)$, where the vertices are the problem variables ($V = X$) and where $S = \{S_1, ..., S_r\}$ is a set of subsets of $V$, called hyperedges, which represent the scopes of the functions in the problem ($S_i = scope(f_i)$). The **primal graph** of a hypergraph $H = (V, S)$ is an undirected graph $G = (V, E)$ such that there is an edge $(u, v) \in E$*

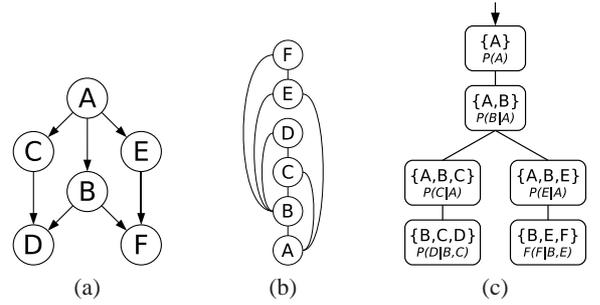

Figure 1: Example belief network, its triangulated primal graph along ordering $d = A, B, C, D, E, F$, and the corresponding bucket tree decomposition.

*for any two vertices $u, v \in V$ that appear in the same hyperedge (namely, there exists $S_i$, s.t., $u, v \in S_i$). The **dual graph** of a hypergraph $H = (V, S)$ is an undirected graph $G = (S, E)$ that has a vertex for each hyperedge, and there is an edge $(S_i, S_j) \in E$ when the corresponding hyperedges share a vertex ($S_i \cap S_j \neq \emptyset$).*

DEFINITION **2.2** *A hypergraph is a **hypertree**, also called **acyclic hypergraph**, if and only if its dual graph has an edge subgraph that is a tree, such that all the nodes in the dual graph that contain a common variable form a connected subgraph.*

It is well-known that problems whose underlying graph have tree structure can be solved efficiently [Pearl, 1988]. If this is not the case, we aim to transform the problem into an equivalent one that exhibits tree structure [Lauritzen and Spiegelhalter, 1988, Dechter and Pearl, 1989, Kask et al., 2005]. Intuitively, we do this by grouping variables and the functions over them into clusters that can be arranged as a tree:

DEFINITION **2.3** *Let $X, D, F$ be the variables, domains and functions of a reasoning problem $\mathcal{P}$. A **tree decomposition** of $\mathcal{P}$ is a triple $\langle T, \chi, \psi \rangle$, where $T = (V, E)$ is a tree and $\chi$ and $\psi$ are labeling functions that associate with each vertex $v \in V$ two sets, $\chi(v) \subseteq X$ and $\psi(v) \subseteq F$, that satisfy the following conditions:*

1. *For each $f_i \in F$, there is at least one vertex $v \in V$ such that $f_i \in \psi(v)$ .*
2. *If $f_i \in \psi(v)$, then $scope(f_i) \subseteq \chi(v)$ .*
3. *For each variable $X_i \in X$, the set $\{v \in V | X_i \in \chi(v)\}$ induces a connected subtree of $T$. This is also called the running intersection property.*

*The **tree width** of a tree decomposition $\langle T, \chi, \psi \rangle$ is $w = \max_v |\chi(v)| - 1$ . The tree width $w^*$ of $\mathcal{P}$ is the minimum tree width over all its tree decompositions.*

The problem of finding the tree decomposition of minimal tree width is known to be NP-complete. To obtain tree decompositions in practice, one can apply a triangulation algorithm to the problem's primal graph along an ordering and then construct the *bucket tree* by extracting and connecting the cliques for each variable, as described for instance in [Pearl, 1988]. The ordering to use as the basis for the triangulation algorithm is often computed heuristically.

**Example 2.1** *Assume the belief network in Figure 1(a) over variables $X = \{A, B, C, D, E, F\}$ is given. We pick the ordering $d = A, B, C, D, E, F$ and triangulate the primal graph as shown in Figure 1(b). If we extract each variable's* bucket – *the variable and its earlier neighbors – we obtain the tree decomposition shown in Figure 1(c), the* bucket tree decomposition.

## 2.2 SOLVING REASONING PROBLEMS

Two principal methods exist to solve reasoning problems, search (e.g., depth-first branch-and-bound, best-first search) and inference (e.g., variable elimination, join-tree clustering). Both can be shown to be time and space exponential in the problem instance's tree width [Lauritzen and Spiegelhalter, 1988, Dechter and Pearl, 1989, Kask et al., 2005, Dechter and Mateescu, 2007], with a dominant factor of $k^w$, where $k$ denotes the maximum domain of the problem variables.

### 2.2.1 Search

Search-based algorithms traverse the problem *search space*. Given a variable ordering $d$, the simplest way to perform search is to instantiate variables one at a time. This will define a search tree, where each node represents a state in the space of partial assignments. Leaf nodes signify either full solutions or dead ends. Standard depth-first algorithms typically have time complexity exponential in the number of variables and require linear space. If memory is available, one can apply caching to traversed nodes and retrieve their values when "similar" nodes are encountered.

These elementary search spaces, however, don't fully capture the structure of the underlying graphical model. Introducing *AND* nodes into the search space can exploit independence of subproblems by effectively conditioning on values, thus avoiding redundant computation [Dechter and Mateescu, 2007]. Since the size of the *AND/OR search tree* may be exponentially smaller than the traditional OR search one, any algorithm exploring the AND/OR space enjoys a better computational bound.

**Example 2.2** *Figure 2 depicts the AND/OR search tree for the problem introduced in Example 2.1 if we assume binary variable domains and the ordering $d = A, B, C, D, E, F$. The AND nodes for variable $B$ each have two OR children, expressing that at this point the problem decomposes into independent subproblems, rooted at $C$ and $E$, respectively.*

We can equally apply caching techniques to an algorithm exploring the AND/OR search tree. As a result this algorithm will effectively explore the *AND/OR search graph*. With caching, identical subproblems are recognized based on their context, which is a graphical model parameter that denotes the part of the search tree above that is relevant to the subproblem below.

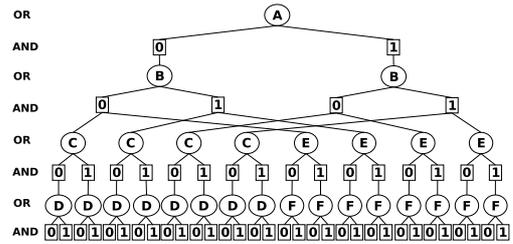

Figure 2: The AND/OR search tree for the example problem, assuming binary variables.

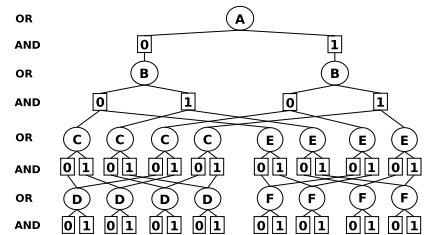

Figure 3: The AND/OR search graph explored by AND/OR search with caching.

Caching will avoid redundant computations, thus reducing time complexity, at the cost of increased memory requirements. By varying the maximum size of contexts to cache on, this tradeoff can be fine-tuned. Assuming full caching, search has been shown to exhibit both time and space complexity exponential in the problem's tree width.

**Example 2.3** *If we extend AND/OR search with caching, given the problem in Example 2.1 it will explore the AND/OR search graph shown in Figure 3. Note how the child nodes of variables $C$ and $E$ are merged. This expresses the fact that in the example the subproblems rooted at $D$ and $F$, as children of $C$ and $E$, respectively, are independent of the value of $A$ further up.*

## 2.3 EXPLOITING DETERMINISM

In practice, however, problem instances across many domains will exhibit a significant degree of determinism (e.g., disallowed tuples in constraint problems, zero probability entries in belief networks). Search algorithms detect the resulting inconsistencies early in the search process and prune the respective portion of the search space. This can lead to significant savings in running time, but is not reflected in the standard worst-case bounds described above.

To exploit determinism in the context of variable elimination, the concept of (generalized) hypertree decompositions has been introduced for constraint networks in

[Gottlob et al., 2000]. As a subclass of tree decompositions, it was shown that it provides a stronger indicator of tractability than the tree width.

DEFINITION **2.4** *Let* $\mathcal{T} = \langle T, \chi, \psi \rangle$, *where* $T = (V, E)$ *be a tree decomposition of a reasoning problem* $\mathcal{P}$ *over a graphical model with variables* $X$, *their domains* $D$ *and functions* $F$. $\mathcal{T}$ *is a* **hypertree decomposition** *of* $\mathcal{P}$ *if the following additional condition is satisfied:*

4. *For each* $v \in V$, $\chi(v) \subseteq \bigcup_{f_j \in \psi(v)} scope(f_j)$.

*The* **hypertree width** *of a hypertree decomposition is* $hw = \max_v |\psi(v)|$. *The hypertree width* $hw^*$ *of* $\mathcal{P}$ *is the minimum hypertree width over all its hypertree decompositions.*

To analyze the complexity of algorithms operating on hypertree decompositions, we introduce the notion of *tightness* of a function or relation:

DEFINITION **2.5** *The* **tightness** $t$ *of a function* $f$ *is the number of relevant tuples (e.g., allowed tuples in constraints, nonzero entries in conditional probability tables).*

The motivation behind this is to store and process the function in a "compressed" form, with only the $t$ relevant tuples. Given a hypertree decomposition, one can then modify an inference algorithm to make use of these compact representations when computing the messages to be passed.

In [Gottlob et al., 2000] the complexity of processing a hypertree decomposition for solving a constraint satisfaction problem is shown to be exponential in $hw^*$, with a dominant factor of $t^{hw^*}$. This result was extended in [Kask et al., 2005] to any graphical model that is absorbing relative to 0. (A graphical model is absorbing relative to a 0 element if its combination operator has the property that $x \bigotimes 0 = 0 \; \forall x$; for example, multiplication has this property while summation does not.)

## 3 HYPERTREE WIDTH BOUNDS FOR INFERENCE

In this section we briefly explore whether the bounds based on hypertree width can provide a practical improvement over the established tree width bounds described above. To that end we recently looked at a selection of over 140 problem instances from various domains [Dechter et al., 2008].

We used the code developed for [Dermaku et al., 2005], which is available online. It generates a tree decomposition along a minfill ordering and extends it to a (generalized) hypertree decomposition by applying a greedy heuristic.

We used the lowest tree width $w$ and hypertree width $hw$ out of 20 runs as a basis for our investigation: For every problem instance, we compute the dominant factors of the two worst-case complexity bounds, i.e., $k^w$ and $t^{hw}$, where $k$ denotes the maximum domain size and $t$ the maximum function tightness in the problem instance. In order for the hypertree decomposition to provide a better bound than the tree decomposition, $t^{hw}$ should be significantly smaller than $k^w$.

Intuitively, it is clear that $t \in \mathcal{O}(k^r)$, where $r$ is the maximum function arity of the problem. Hence we should expect that only when the function table contains many irrelevant values (e.g., zeros in probability tables), the hypertree width bound can be superior.

And indeed, out of all the instances we evaluated, only for five of them was $t^{hw}$ less than $k^w$, whereas it was orders of magnitude worse on almost all of the remaining problems. Looking at the instances in more detail, it becomes evident that in most of them the functions are not sufficiently tight and often have highly intersecting scopes, which renders the hypertree width bound ineffective.

## 4 SEARCH SPACE ESTIMATION

Even though the results in [Dechter et al., 2008] suggest that complexity bounds based on hypertree width are often not competitive in practice, the idea of exploiting determinism remains promising. Furthermore, while all considerations in section 3 were targeted at inference algorithms, we are in particular interested in estimating how determinism in a problem will impact the size of the search space discussed in Section 2.2.1, since search is a widespread method in practice. To that end, we will aim to upper bound the size of the AND/OR context minimal search graph, which is explored by AND/OR search augmented with caching, as described in Section 2.

### 4.1 TREE DECOMPOSITION CORRESPONDENCE

Assume that a variable ordering $d = x_1, \ldots, x_n$ is fixed and that search instantiates variables first to last (while inference would proceed last to first). We note that the way the search space is decomposed by AND/OR search with caching can be represented by the bucket tree decomposition along the same ordering. For details we refer to [Mateescu and Dechter, 2005], to illustrate we revisit our previous example:

**Example 4.1** *Consider the bucket tree decomposition in Figure 1(c) and the AND/OR search graph in Figure 3. It is easy to see that the decomposition clusters can be related to the "layers" of the search graph, i.e., the nodes associated with a variable and its values, as shown in Figure 4. For instance, the cluster* $\{B, C, D\}$ *represents the search layer for variable* $D$ *and the fact that the subproblem only depends on* $B$ *and* $C$ *– and not* $A$.

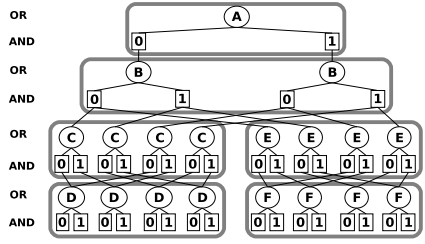

Figure 4: The AND/OR search graph with clusters corresponding to the bucket tree decomposition in Figure 1(c).

Based on this observation, we can also partition the search space into "clusters", according to the corresponding bucket tree decomposition. Our approach of upper bounding the size of the entire search space will then be to bound the portion of the search space in each cluster, subsequently summing over the clusters (for simplicity we consider only the AND nodes of the search space, since OR nodes are actually not implemented as such in practice). We note that a cluster in a bucket tree decomposition will always correspond to exactly one variable's layer in the AND/OR search graph (due to the way it is constructed).

### 4.2 BOUNDING CLUSTER SIZE

A straightforward upper bound is obtained by multiplying the domain sizes of all variables in the cluster [Kjærulff, 1990]. If the set of clusters is given by $C = \{C_1, \ldots, C_n\}, C_k \subseteq X$, summing over clusters gives

$$twb := \sum_{k=1}^{n} \prod_{x_i \in C_k} |D_i| .$$

Note that this is very closely related to the worst case complexity, since the tree width is just the maximum number of variables in any cluster of the decomposition. This, however, does not take determinism into account.

If the bucket tree decomposition we are working with is also a hypertree decomposition (meaning all variables in a cluster are covered by the functions in that cluster), we can take the product $\prod_i t_i$ as an upper bound to the number of nodes in that cluster, where $t_i$ is the tightness of the $i$-th function in it. (This is closely related to the complexity bound on hypertree decompositions as outlined above.)

Taking this one step further, if the bucket tree decomposition at hand does not satisfy the additional hypertree decomposition condition, we can compute the product over the tightness of each function in the cluster and multiply this by the domain sizes of the uncovered variables, to account for the lack of information about these variables.

However, since the scopes of the different functions in a cluster can overlap, this tightness based bound will typically be worse than the simple "product of domain sizes" (as we already observed). Consequentially, we use the latter as a starting point and exploit tight functions to improve

**Algorithm** *GreedyCovering*
**Input:** Cluster $C_k$ containing variables $X_k$ and functions $F_k$, with $x_i \in X_k$ having domain size $d_i$ and $f_j \in F_k$ having tightness $t_j$
**Output:** A subset of $F_k$ (forming a partial covering of $X_k$)
**Init:** $Uncov = X_k$, $Covering = \emptyset$
(1) Find $j^*$ that minimizes $r_j = t_j / \prod_{x_k \in I_j} d_k$, where
    $I_j = Uncov \cap scope(f_j)$.
(2) If $r_{j^*} \geq 1$, terminate with current covering.
(3) Add $f_{j^*}$ to $Covering$ and set $Uncov := Uncov \setminus scope(f_{j^*})$.
(4) If $Uncov = \emptyset$, terminate with current covering.
(5) Goto (1).

Figure 5: Greedy covering algorithm for a single cluster.

upon it, therefore combining the concept of tree and hypertree decompositions.

In essence this can be seen as a weighted variant of the well-known SET COVER problem, where one aims to cover a set or vertices by as few as possible subsets from a set of given subsets of the variables. The problem is generally NP-complete, but simple greedy approximations exist [Johnson, 1973], which give rise to our method:

Start with an empty covering (if we assume dummy unary functions over uncovered variables, this is equivalent to the bound $twb$). Then, for each function $f_j$ in the cluster, compute the *coverage ratio* $r_j$ as follows: Divide the function's tightness $t_j$ by the product over the domain sizes of the variables that have not yet been covered and are in the scope of $f_j$. Pick the function for which the coverage ratio is the lowest and add it to the covering. Repeat this for as long as we can still find a function with a coverage ratio less than 1. The algorithm is given in Figure 5.

It will produce a set of functions as the covering, but might leave some variables uncovered. As before, we can multiply the tightness of the functions in the covering and the domain sizes of the uncovered variables to obtain an upper bound on the number of nodes in the cluster.

It is worth noting that we are not limited to functions from the decomposition cluster in question, but we can include any function from the clusters higher up in the rooted tree decomposition (since their scope will have been fully instantiated at this point of the search).

**Proposition 1** *Executing algorithm* GreedyCovering *for each cluster of the bucket tree decomposition and summing up, we obtain an upper bound on the number of nodes in the AND/OR search graph, which we denote:*

$$hwb := \sum_{k=1}^{n} \left( \prod_{f_j \in Cov(C_k)} t_j \cdot \prod_{x_i \in C_k \setminus Cov(C_k)} |D_i| \right),$$

*where $Cov(C_k)$ is the set of functions returned by the algorithm* GreedyCovering *for cluster $C_k$.*

THEOREM **4.2** *Given a reasoning problem with $n$ variables and $m$ relations with maximal tightness $t$. If the bucket tree decomposition along a given variable ordering $d$ has tree width $w$, the complexity of computing the bound $hwb$ is $\mathcal{O}(n \cdot w \cdot m)$ time-wise and $\mathcal{O}(m)$ space-wise.*

**Proof.** The time complexity of algorithm *GreedyCovering* is linear both on the number of variables in the cluster as well as in the number of functions considered, i.e., worst-case $\mathcal{O}(w \cdot m)$. Keeping track of the coverage ratio of each function requires $\mathcal{O}(m)$ space. Iterating and summing over all $n$ clusters results in the stated asymptotic bounds. □

**Example 4.3** *Assume we have a cluster containing 3 variables $X$, $Y$, $Z$ with domain sizes $d_X = 4$, $d_Y = 4$, and $d_Z = 3$, as well as 2 functions $f_1(X,Y)$ and $f_2(Y,Z)$ with tightness $t_1 = 9$ and $t_2 = 11$. The $twb$ bound on the number of nodes in this cluster is $d_X \cdot d_Y \cdot d_Z = 48$. In the first iteration of the greedy covering algorithm for $hwb$ the gain ratios of $f_1$ and $f_2$ will be computed as $\frac{9}{16}$ and $\frac{11}{12}$, respectively. Therefore $f_1$ will be added to the covering, leaving only $Z$ uncovered. The next gain ratio of $f_2$ will be computed as $\frac{11}{3}$, which is greater than 1. Therefore the algorithm terminates and $f_2$ will not be part of the covering. The $hwb$ bound for this cluster will then be $t_1 \cdot d_Z = 9 \cdot 3 = 27$.*

## 5 EXPERIMENTAL RESULTS

For empirical evaluation we return to the problem instances that were described in [Dechter et al., 2008] (see also Section 3). These comprise 112 belief networks from areas such as coding networks, dynamic Bayesian networks, genetic linkage instances and CPCS medical diagnosis networks. We also evaluated 30 weighted constraint network instances. All problem instances are available online[1].

For every instance, we report the number of variables $n$, the maximum variable domain size $k$, the maximum function arity $r$, and the average tightness ratio $tr$ (defined as the average percentage of relevant tuples in a full function table).

On each problem instance we run our bounding method along 100 different minfill orderings (with random tie breaking) and record the lowest bound, with $w$ as the tree width of the bucket tree decomposition. For every instance we then compute the asymptotic worst-case bound for the search space size, which is $n \cdot k^{w+1}$ (cf. [Dechter and Mateescu, 2007], adapted for AND nodes).

Consulting the bucket tree decomposition for a more fine-grained analysis (still without considering determinism) gives the bound $twb$. We then apply our covering heuristic to exploit determinism and obtain the bound $hwb$. (Note that these three bounds can produce very large integers, therefore we report them as their $\log_{10}$ in Table 1.) Even for the larger problem instances this bound computation takes only a few hundred milliseconds on a 2.66 GHz CPU.

[1] http://graphmod.ics.uci.edu/

### 5.1 BOUND IMPROVEMENTS

If we compare the values for $n \cdot k^{w+1}$ and $twb$, we can see that the bound improves for every problem instance by at least one order of magnitude, but often more than that (recall that the table shows $\log_{10}$ of the bounds). For most pedigree genetic linkage instances, for example, the reduction is ten orders of magnitude or more. Similar results hold for the dynamic Bayesian networks we tested on, with many orders of magnitude improvement.

It seems that problems with a higher number of variables benefit the most from the fine-grained analysis. This makes sense if one considers the fact that the worst-case bound will greatly overestimate the size of almost all clusters, since in practice the tree decomposition contains only very few clusters of full tree width.

If we try to exploit determinism by going from the $twb$ to $hwb$ bound, there is, just looking at the problem parameters, no obvious indicator for when the bound will improve: On pedigree instances, for example, the decrease is not very significant, although these problems exihibit some determinism. On digital circuits, on the other hand, with an average of 50% determinism, the bound improves another 3 to 4 orders of magnitude over $twb$.

On almost all weighted CSP instances we were able to lower the bound by exploiting determinism, often by orders of magnitude. For example, on the satellite scheduling problem 408b the $twb$ bound of $83,206,198,094$ decreases to a $hwb$ value of $248,197$. A significantly tighter bound is also achieved on randomly generated $k$-trees where the probability tables were forced to exihibit determinism, with for example $twb = 29,983,742$ decreasing to $hwb = 1,528$ on problem BN_107.

The crucial point here seems to be at which point during the search functions with high determinism will have their scope fully instantiated, i.e., at which point they become available to our covering heuristic. This is not predictable by only looking at the instance parameters but will require a more detailed look at the guiding bucket tree decomposition instead.

### 5.2 BOUND EVALUATION

Most problem instances are too big to compute the exact size of the context minimal AND/OR search space, which would be equivalent to solving the problem (for solution counting or computing $P(e)$). But for some of the smaller instances this is actually feasible, which gives us the option of testing how tight our bound is.

| instance | $n$ | $k$ | $r$ | $tr$ | $w$ | $\log_{10}$ $nk^{w+1}$ | $twb$ | $hwb$ | instance | $n$ | $k$ | $r$ | $tr$ | $w$ | $\log_{10}$ $nk^{w+1}$ | $twb$ | $hwb$ |
|---|---|---|---|---|---|---|---|---|---|---|---|---|---|---|---|---|---|
| Grid networks | | | | | | | | | Coding networks | | | | | | | | |
| 90-10-1 | 100 | 2 | 3 | 0.58 | 9 | 5.01 | 4.06 | 3.92 | BN_126 | 512 | 2 | 5 | 0.87 | 53 | 18.96 | 16.81 | 16.81 |
| 90-14-1 | 196 | 2 | 3 | 0.55 | 15 | 7.11 | 5.63 | 5.58 | BN_127 | 512 | 2 | 5 | 0.88 | 57 | 20.17 | 17.86 | 17.86 |
| 90-16-1 | 256 | 2 | 3 | 0.56 | 17 | 7.83 | 6.25 | 6.15 | BN_128 | 512 | 2 | 5 | 0.88 | 48 | 17.46 | 15.48 | 15.48 |
| 90-24-1 | 576 | 2 | 3 | 0.55 | 15 | 7.58 | 5.91 | 5.76 | BN_129 | 512 | 2 | 5 | 0.88 | 52 | 18.66 | 16.66 | 16.66 |
| 90-24-1e20 | 576 | 2 | 3 | 0.55 | 31 | 12.39 | 10.12 | 10.12 | BN_130 | 512 | 2 | 5 | 0.88 | 54 | 19.27 | 16.98 | 16.98 |
| 90-26-1e40 | 676 | 2 | 3 | 0.55 | 29 | 11.86 | 9.88 | 9.83 | BN_131 | 512 | 2 | 5 | 0.87 | 48 | 17.46 | 15.42 | 15.42 |
| 90-30-1e60 | 900 | 2 | 3 | 0.55 | 37 | 14.39 | 11.96 | 11.95 | BN_132 | 512 | 2 | 5 | 0.88 | 49 | 17.76 | 16.00 | 16.00 |
| 90-34-1e80 | 1156 | 2 | 3 | 0.56 | 39 | 15.10 | 12.66 | 12.53 | BN_133 | 512 | 2 | 5 | 0.87 | 54 | 19.27 | 17.26 | 17.26 |
| 90-38-1e120 | 1444 | 2 | 3 | 0.55 | 43 | 16.40 | 13.97 | 13.69 | BN_134 | 512 | 2 | 5 | 0.87 | 52 | 18.66 | 16.41 | 16.41 |
| Dynamic Bayesian Networks | | | | | | | | | CPCS medical diagnosis | | | | | | | | |
| BN_21 | 2843 | 91 | 4 | 0.49 | 6 | 17.17 | 8.50 | 7.82 | cpcs54 | 54 | 2 | 10 | 1.00 | 12 | 5.65 | 4.68 | 4.68 |
| BN_23 | 2425 | 91 | 4 | 0.47 | 4 | 13.18 | 7.37 | 6.59 | cpcs179 | 179 | 4 | 9 | 1.00 | 7 | 7.07 | 5.04 | 5.04 |
| BN_25 | 1819 | 91 | 4 | 0.53 | 4 | 13.06 | 7.17 | 6.77 | cpcs360b | 360 | 2 | 12 | 1.00 | 16 | 7.67 | 5.50 | 5.50 |
| BN_27 | 3025 | 5 | 7 | 1.00 | 9 | 10.47 | 7.33 | 7.33 | cpcs422b | 422 | 2 | 18 | 0.99 | 22 | 9.55 | 7.51 | 7.51 |
| BN_29 | 24 | 10 | 6 | 1.00 | 5 | 7.38 | 6.16 | 6.16 | Genetic linkage | | | | | | | | |
| Grid networks | | | | | | | | | pedigree1 | 334 | 4 | 5 | 0.79 | 15 | 12.06 | 6.88 | 6.85 |
| BN_31 | 1156 | 2 | 3 | 0.56 | 35 | 13.90 | 11.53 | 11.47 | pedigree18 | 1184 | 5 | 5 | 0.81 | 20 | 17.75 | 7.58 | 7.58 |
| BN_33 | 1444 | 2 | 3 | 0.56 | 37 | 14.60 | 12.42 | 12.26 | pedigree20 | 437 | 5 | 4 | 0.79 | 22 | 18.72 | 9.42 | 9.20 |
| BN_35 | 1444 | 2 | 3 | 0.55 | 38 | 14.90 | 12.51 | 12.25 | pedigree23 | 402 | 5 | 4 | 0.80 | 27 | 22.18 | 11.73 | 10.85 |
| BN_37 | 1444 | 2 | 3 | 0.55 | 40 | 15.50 | 13.01 | 12.99 | pedigree25 | 1289 | 5 | 4 | 0.83 | 24 | 20.58 | 9.18 | 9.18 |
| BN_39 | 1444 | 2 | 3 | 0.56 | 38 | 14.90 | 12.63 | 12.57 | pedigree30 | 1289 | 5 | 5 | 0.82 | 21 | 18.49 | 7.95 | 7.95 |
| BN_41 | 1444 | 2 | 3 | 0.56 | 40 | 15.50 | 13.09 | 12.99 | pedigree33 | 798 | 4 | 5 | 0.81 | 30 | 21.57 | 11.37 | 10.20 |
| Digital circuits | | | | | | | | | pedigree37 | 1032 | 5 | 4 | 0.82 | 21 | 18.39 | 10.84 | 10.74 |
| BN_48 | 661 | 2 | 5 | 0.51 | 43 | 16.07 | 13.79 | 10.44 | pedigree38 | 724 | 5 | 4 | 0.78 | 16 | 14.74 | 10.72 | 10.52 |
| BN_50 | 661 | 2 | 5 | 0.51 | 43 | 16.07 | 13.79 | 10.69 | pedigree39 | 1272 | 5 | 4 | 0.85 | 20 | 17.78 | 8.21 | 8.12 |
| BN_52 | 661 | 2 | 5 | 0.51 | 41 | 15.46 | 13.39 | 9.86 | pedigree42 | 448 | 5 | 4 | 0.79 | 23 | 19.43 | 10.66 | 10.14 |
| BN_54 | 561 | 2 | 5 | 0.53 | 48 | 17.50 | 15.43 | 13.05 | pedigree50 | 514 | 6 | 4 | 0.77 | 18 | 17.50 | 11.57 | 11.53 |
| BN_56 | 561 | 2 | 5 | 0.53 | 51 | 18.40 | 16.19 | 14.04 | pedigree7 | 1068 | 4 | 4 | 0.83 | 32 | 22.90 | 11.94 | 11.71 |
| BN_58 | 561 | 2 | 5 | 0.53 | 50 | 18.10 | 15.72 | 13.09 | pedigree9 | 1118 | 7 | 4 | 0.79 | 26 | 25.87 | 9.88 | 9.86 |
| BN_60 | 540 | 2 | 5 | 0.53 | 55 | 19.59 | 17.35 | 14.43 | pedigree13 | 1077 | 3 | 4 | 0.83 | 34 | 19.73 | 12.04 | 11.99 |
| BN_62 | 667 | 2 | 5 | 0.51 | 42 | 15.77 | 13.93 | 10.73 | pedigree19 | 793 | 5 | 5 | 0.78 | 23 | 19.67 | 10.00 | 9.99 |
| BN_64 | 540 | 2 | 5 | 0.53 | 50 | 18.08 | 15.86 | 14.07 | pedigree31 | 1183 | 5 | 5 | 0.81 | 30 | 24.74 | 11.77 | 11.77 |
| BN_66 | 440 | 2 | 5 | 0.55 | 59 | 20.71 | 18.82 | 16.15 | pedigree34 | 1160 | 5 | 4 | 0.83 | 32 | 26.13 | 12.16 | 12.16 |
| BN_68 | 440 | 2 | 5 | 0.55 | 57 | 20.10 | 18.08 | 15.48 | pedigree40 | 1030 | 7 | 5 | 0.80 | 29 | 28.37 | 12.38 | 12.38 |
| CPCS medical diagnosis | | | | | | | | | pedigree41 | 1062 | 5 | 5 | 0.80 | 32 | 26.09 | 12.26 | 12.05 |
| BN_79 | 54 | 2 | 10 | 1.00 | 10 | 5.04 | 4.23 | 4.23 | pedigree44 | 811 | 4 | 5 | 0.80 | 26 | 19.16 | 10.13 | 9.98 |
| BN_81 | 360 | 2 | 12 | 0.93 | 16 | 7.67 | 5.75 | 5.74 | pedigree51 | 1152 | 5 | 4 | 0.82 | 38 | 30.32 | 12.89 | 12.84 |
| BN_83 | 360 | 2 | 12 | 0.97 | 18 | 8.28 | 6.31 | 6.31 | Digital circuits | | | | | | | | |
| BN_85 | 360 | 2 | 12 | 0.99 | 19 | 8.58 | 6.61 | 6.61 | c432.isc | 432 | 2 | 10 | 0.54 | 20 | 8.96 | 6.96 | 5.43 |
| BN_87 | 422 | 2 | 18 | 0.98 | 21 | 9.25 | 7.36 | 7.36 | c499.isc | 499 | 2 | 6 | 0.54 | 19 | 8.72 | 6.98 | 5.19 |
| BN_89 | 422 | 2 | 18 | 0.97 | 17 | 8.04 | 6.33 | 6.33 | s386.scan | 172 | 2 | 5 | 0.54 | 16 | 7.35 | 5.71 | 4.61 |
| BN_91 | 422 | 2 | 18 | 0.98 | 21 | 9.25 | 7.31 | 7.31 | s953.scan | 440 | 2 | 5 | 0.54 | 26 | 10.77 | 8.85 | 6.50 |
| BN_93 | 422 | 2 | 18 | 0.97 | 20 | 8.95 | 6.99 | 6.99 | Various networks | | | | | | | | |
| Randomly generated belief networks | | | | | | | | | Barley | 48 | 67 | 5 | 0.98 | 7 | 16.29 | 7.26 | 7.26 |
| BN_95 | 53 | 3 | 4 | 1.00 | 15 | 9.36 | 7.01 | 7.01 | Diabetes | 413 | 21 | 3 | 0.45 | 4 | 9.23 | 7.10 | 6.83 |
| BN_97 | 54 | 3 | 4 | 1.00 | 15 | 9.37 | 7.13 | 7.13 | hailfinder | 56 | 11 | 5 | 0.84 | 4 | 6.96 | 4.01 | 3.78 |
| BN_99 | 57 | 3 | 4 | 1.00 | 16 | 9.87 | 7.70 | 7.70 | insurance | 27 | 5 | 4 | 0.84 | 6 | 6.32 | 4.50 | 4.45 |
| BN_101 | 58 | 3 | 4 | 1.00 | 15 | 9.40 | 7.00 | 7.00 | Mildew | 35 | 100 | 4 | 0.62 | 4 | 11.54 | 6.57 | 6.03 |
| BN_103 | 76 | 3 | 4 | 1.00 | 17 | 10.47 | 7.43 | 7.43 | Munin1 | 189 | 21 | 4 | 0.49 | 11 | 18.14 | 8.31 | 8.15 |
| Randomly generate partial $k$-trees with forced determinism | | | | | | | | | Munin2 | 1003 | 21 | 4 | 0.48 | 8 | 14.90 | 6.86 | 6.70 |
| BN_105 | 50 | 2 | 21 | 0.60 | 18 | 7.42 | 6.39 | 2.54 | Munin3 | 1044 | 21 | 4 | 0.47 | 9 | 16.24 | 6.95 | 6.72 |
| BN_107 | 50 | 2 | 21 | 0.59 | 21 | 8.32 | 7.48 | 3.18 | Munin4 | 1041 | 21 | 4 | 0.47 | 9 | 16.24 | 7.54 | 7.27 |
| BN_109 | 50 | 2 | 20 | 0.62 | 20 | 8.02 | 7.12 | 3.49 | Pigs | 441 | 3 | 3 | 0.70 | 10 | 7.89 | 5.90 | 5.89 |
| BN_111 | 50 | 2 | 20 | 0.63 | 19 | 7.72 | 6.92 | 3.31 | Water | 32 | 4 | 6 | 0.58 | 10 | 8.13 | 6.66 | 6.20 |
| BN_113 | 50 | 2 | 21 | 0.62 | 21 | 7.28 | 7.28 | 3.38 | Genetic linkage | | | | | | | | |
| Randomly generate partial $k$-trees without forced determinism | | | | | | | | | fileEA0 | 381 | 4 | 4 | 0.81 | 7 | 7.40 | 3.92 | 3.68 |
| BN_115 | 50 | 2 | 19 | 1.00 | 20 | 8.02 | 7.07 | 7.07 | fileEA1 | 836 | 5 | 4 | 0.82 | 11 | 11.31 | 4.68 | 4.16 |
| BN_117 | 50 | 2 | 20 | 1.00 | 18 | 7.42 | 6.43 | 6.43 | fileEA2 | 979 | 5 | 4 | 0.82 | 11 | 11.38 | 4.86 | 4.48 |
| BN_119 | 50 | 2 | 19 | 1.00 | 19 | 7.72 | 6.60 | 6.60 | fileEA3 | 1122 | 5 | 4 | 0.82 | 13 | 12.84 | 5.19 | 4.70 |
| BN_121 | 50 | 2 | 19 | 1.00 | 19 | 7.72 | 6.75 | 6.75 | fileEA4 | 1231 | 5 | 4 | 0.82 | 13 | 12.88 | 5.15 | 4.71 |
| BN_123 | 50 | 2 | 20 | 1.00 | 18 | 7.42 | 6.46 | 6.46 | fileEA5 | 1515 | 5 | 4 | 0.82 | 12 | 12.27 | 5.27 | 4.94 |
| BN_125 | 50 | 2 | 18 | 1.00 | 19 | 7.72 | 6.70 | 6.70 | fileEA6 | 1816 | 5 | 4 | 0.82 | 14 | 13.74 | 5.85 | 5.36 |
| Digital circuits (WCSP) | | | | | | | | | Satellite scheduling (WCSP) | | | | | | | | |
| c432 | 432 | 2 | 10 | 0.61 | 27 | 11.06 | 8.90 | 8.90 | 29 | 82 | 4 | 2 | 0.74 | 14 | 10.94 | 8.41 | 6.01 |
| c499 | 499 | 2 | 6 | 0.63 | 23 | 9.92 | 8.03 | 7.25 | 42b | 190 | 4 | 2 | 0.78 | 18 | 13.72 | 10.46 | 7.19 |
| c880 | 880 | 2 | 5 | 0.64 | 24 | 10.47 | 7.93 | 6.90 | 54 | 67 | 4 | 3 | 0.75 | 11 | 9.05 | 6.33 | 4.49 |
| s1196 | 561 | 2 | 5 | 0.82 | 51 | 18.40 | 16.37 | 13.91 | 404 | 100 | 4 | 3 | 0.74 | 19 | 14.04 | 7.65 | 3.83 |
| s1238 | 540 | 2 | 5 | 0.87 | 54 | 19.29 | 17.04 | 14.82 | 408b | 200 | 4 | 2 | 0.75 | 24 | 17.35 | 10.58 | 5.40 |
| s1423 | 748 | 2 | 5 | 0.78 | 22 | 9.80 | 7.54 | 7.54 | 503 | 143 | 4 | 3 | 0.76 | 9 | 8.18 | 5.77 | 4.22 |
| s1488 | 667 | 2 | 5 | 0.90 | 44 | 16.37 | 14.34 | 10.99 | 505b | 240 | 4 | 2 | 0.75 | 16 | 12.62 | 8.87 | 6.67 |
| s1494 | 661 | 2 | 5 | 0.91 | 44 | 16.37 | 14.34 | 10.76 | Radio frequency assignment (WCSP) | | | | | | | | |
| s386 | 172 | 2 | 5 | 0.82 | 18 | 7.96 | 6.46 | 4.99 | C6-sub0 | 16 | 44 | 2 | 0.31 | 7 | 14.35 | 12.99 | 10.20 |
| s953 | 440 | 2 | 5 | 0.80 | 62 | 21.61 | 19.69 | 16.30 | C6-sub1-24 | 14 | 24 | 2 | 0.26 | 9 | 14.95 | 14.12 | 9.67 |
| Mastermind puzzle game (WCSP) | | | | | | | | | C6-sub1 | 14 | 44 | 2 | 0.24 | 9 | 17.58 | 16.75 | 11.76 |
| 03_08_03 | 1220 | 2 | 3 | 0.85 | 20 | 9.41 | 7.28 | 5.29 | C6-sub2 | 16 | 44 | 2 | 0.24 | 10 | 19.28 | 18.01 | 12.05 |
| 03_08_04 | 2288 | 2 | 3 | 0.87 | 30 | 12.69 | 10.37 | 7.96 | C6-sub3 | 18 | 44 | 2 | 0.26 | 10 | 19.33 | 18.02 | 12.48 |
| 03_08_05 | 3692 | 2 | 3 | 0.88 | 38 | 15.31 | 12.73 | 8.77 | C6-sub4-20 | 22 | 20 | 2 | 0.34 | 11 | 16.95 | 15.94 | 11.50 |
| 04_08_03 | 1418 | 2 | 3 | 0.85 | 24 | 10.68 | 8.57 | 6.68 | C6-sub4 | 22 | 44 | 2 | 0.30 | 11 | 21.06 | 19.78 | 15.30 |
| 04_08_04 | 2616 | 2 | 3 | 0.88 | 36 | 14.56 | 11.99 | 9.15 | | | | | | | | | |
| 10_08_03 | 2606 | 2 | 3 | 0.88 | 48 | 18.17 | 15.80 | 13.60 | | | | | | | | | |

Table 1: Results for experiments on 112 Bayesian networks and 30 weighted CSP instances.

| instance | $nk^{w+1}$ | $twb$ | $hwb$ | $\#cm$ |
|---|---|---|---|---|
| 90-10-1 | 204,800 | 14,154 | 11,908 | 11,519 |
| 90-14-1 | 25,690,112 | 804,822 | 689,786 | 683,823 |
| 90-16-1 | 134,217,728 | 2,637,878 | 2,335,466 | 188,625 |
| 90-24-1 | 150,994,944 | 1,286,726 | 1,115,509 | 52,802 |
| cpcs54 | 442,368 | 48,842 | 48,842 | 48,842 |
| cpcs179 | 11,730,944 | 110,560 | 110,512 | 110,512 |
| cpcs360b | 47,185,920 | 319,724 | 319,623 | 319,623 |
| c432.isc | 1,811,939,328 | 10,793,946 | 685,001 | 683,823 |
| c499.isc | 1,046,478,848 | 12,089,118 | 189,637 | 188,625 |
| s386.scan | 45,088,768 | 802,526 | 94,830 | 52,802 |
| s953.scan | 472,446,402,560 | 2,685,782,044 | 4,547,508 | 236,430 |
| fileEA0 | 24,969,216 | 9,454 | 5,316 | 4,774 |
| fileEA1 | 1,020,507,812,500 | 63,520 | 18,444 | 14,057 |
| fileEA2 | 5,975,341,796,875 | 167,630 | 33,851 | 24,203 |
| fileEA3 | 34,240,722,656,250 | 253,170 | 58,147 | 45,052 |
| fileEA4 | 939,178,466,796,875 | 675,230 | 53,214 | 30,868 |
| fileEA5 | 9,246,826,171,875 | 282,454 | 101,825 | 36,146 |
| fileEA6 | 6,927,490,234,375,000 | 2,460,002 | 333,198 | 62,041 |

Table 2: Comparison of bounds to exact search space size $\#cm$.

Table 2 shows the upper bounds $nk^{w+1}$, $twb$, and $hwb$ (this time not in their $\log_{10}$), as well as the exact number of AND nodes in the actual context-minimal search graph. The values in each row were obtained on the same minfill ordering (not neccessarily the one used for Table 1).

For smaller instances the bound we compute turns out to be rather tight (note that CPCS instances exhibit no determinism at all and thus $twb$ and $hwb$ match the size of the search space exactly). As the problems become bigger and their structure more complicated, however, the bound quality deteriorates. It should be interesting to perform this comparison on bigger problem instances, but as of now this is limited by the resources available in current computers.

## 6 CONCLUSIONS & FUTURE WORK

While asymptotic bounds for search algorithms can give a rough idea about problem hardness, it is often desirable to obtain a tighter, more fine-grained bound. As has previously been shown, this can be accomplished by looking at a suitable tree decomposition of the problem's underlying graph structure and the domains of variables in the decomposition clusters. This, however, is blind to determinism, which can greatly prune the search space in practice.

The contribution of this paper is to introduce ideas from the framework of hypertree decompositions into the bounding of the search space. This allows us to exploit determinism in the function specification, but only if it is beneficial to the overall complexity bound.

We demonstrated on a set of 112 belief networks and 30 weighted constraint networks that the proposed scheme is indeed able to further improve the bound on search complexity, in some cases by several orders of magnitude. On a subset of the instances we also showed that the bound can indeed be very tight, although it seems to deteriorate for bigger instances. In this respect we hope to be able to conduct more in-depth comparisons on even bigger problem instances in the future. Note that the ability to bound, sometimes accurately, the size of the search space in linear time is very important, especially for problem instances which are completely unsolvable exactly.

We believe that the current version of our bounding scheme can be further improved by incorporating some form of propagation of information down the bucket tree. Another path we plan to pursue is using approximate counting methods, such as sampling, to compute approximations to the number of solutions in each cluster, which will also approximate the number of nodes. Finally, for optimization tasks and for approximating branch-and-bound and best-first search algorithms we hope to accomplish further tightening of the search space using the cost function itself.

On a higher level, we plan to use the bounds we obtain to guide the selection of static and dynamic variable orderings. We also intend to deploy our scheme for parallelizing search algorithms over a networks of many machines (e.g., grids and clusters).

### Acknowledgments

This work was partially supported by NSF grant IIS-0713118 and NIH grant R01-HG004175-02.